\documentclass[sigconf]{acmart}

\AtBeginDocument{%
  }

\setcopyright{acmlicensed}
\copyrightyear{2018}
\acmYear{2018}
\acmDOI{XXXXXXX.XXXXXXX}
\acmConference[Conference acronym 'XX]{Make sure to enter the correct
  conference title from your rights confirmation emai}{June 03--05,
  2018}{Woodstock, NY}
\acmISBN{978-1-4503-XXXX-X/18/06}

\usepackage{multirow}

\usepackage[utf8]{inputenc}
\usepackage{amsmath,amssymb,amsfonts}
\usepackage{algorithm}
\usepackage{algpseudocode}
\usepackage{array}
\usepackage{listings}
\usepackage{verbatim}

\definecolor{softgreen}{RGB}{15,157,88}

\begin{document}

\title{O1 Embedder: Let Retrievers Think Before Action} 

\author{Ruiran Yan}

\affiliation{%
  \institution{University of Science and Technology of China}
  \city{Hefei}
  \country{China}
}
\email{yanruiran@mail.ustc.edu.cn} 

\author{Zheng Liu} 
\authornote{Corresponding Author.}
\affiliation{
  \institution{Beijing Academy of Artifical Intelligence}
  \city{Beijing}
  \country{China}
}
\email{zhengliu1026@gmail.com}

\author{Defu Lian}
\authornotemark[1]
\affiliation{%
  \institution{University of Science and Technology of China}
  \city{Hefei}
  \country{China}
}
\email{liandefu@ustc.edu.cn}


\begin{abstract} 
The growing power of large language models (LLMs) has revolutionized how people access and utilize information. Notably, the LLMs excel at performing fine-grained data representation, which facilitates precise retrieval of information. They also generate high-quality answers based on external references, enabling the production of useful knowledge. The recent introduction of reasoning models, like OpenAI O1\footnote{https://openai.com/index/introducing-openai-o1-preview/} and DeepSeek R1\footnote{https://github.com/deepseek-ai/DeepSeek-R1}, marks another leap forward, highlighting LLMs' ability to think progressively before delivering final answers. This breakthrough significantly improves the ability to address complex tasks, e.g., coding and math proofs. 

Inspired by this progress, we aim to develop similar capabilities for retrieval models, which hold great promise for tackling critical challenges in the field, including multi-task retrieval, zero-shot retrieval, and tasks requiring intensive reasoning of complex relationships. With this motivation, we propose a novel approach called \textbf{O1 Embedder}, which generates useful thoughts for the input query before making retrieval for the target documents. To realize this objective, we conquer two technical difficulties. First, we design a data synthesis workflow, creating training signals for O1 Embedder by generating initial thoughts from an LLM-expert and subsequently refining them using a retrieval committee. Second, we optimize the training process, enabling a pre-trained model to be jointly fine-tuned to generate retrieval thoughts via behavior cloning and perform dense retrieval through contrastive learning. Our approach is evaluated by comprehensive experiments, where substantial improvements are achieved across 12 popular datasets, spanning both in-domain and out-of-domain scenarios. These results highlight O1 Embedder's remarkable accuracy and generalizability, paving the way for the development of next-generation IR foundation models. 

\end{abstract}



\maketitle

\section{Introduction}
Information retrieval (IR) is fundamental to many important applications, such as search engines and question answering systems~\cite{karpukhin2020dense, kobayashi2000information}. Recently, it draws even higher attention because of its critical role in augmenting large language models (LLMs) with external knowledge~\cite{zhu2023large,10.1145/3589335.3641299,zhao2024dense}, a paradigm known as retrieval-augmented generation (RAG)~\cite{gao2023retrieval, zhang2023retrieve, zhao2024retrieval}. Over the past decade, IR techniques have experienced tremendous progresses. One important breakthrough is made by dense retrieval, where relevant data can be effectively retrieved via vector search. With the popularity of open-source models in this field \cite{neelakantan_text_2022,wang2023improving,xiao_c-pack_2024}, dense retrieval has become a go-to option for realizing retrieval applications in reality. 


However, dense retrieval is still subject to many limitations in this stage. First, existing methods struggle with \textit{zero-shot retrieval} tasks in unseen scenarios, which differ significantly from their source domains. For instance, a well-trained embedding model from general datasets is prone to a limited performance when applied to a specialized problem, like medical or legal case retrieval~\cite{maia201818,li2024automireffectivezeroshotmedical,10.1145/3451964.3451965}. Second, the existing models are insufficient to discriminate \textit{complex relationships}, as they cannot be identified directly from semantic meaning. For example, the retrieval of useful code snippets for a computer program, or the retrieval of evidence to a multi-hop reasoning problem~\cite{li2024coircomprehensivebenchmarkcode, husain2019codesearchnet, yang2018hotpotqa, lee2022generativemultihopretrieval}. 

The above challenges can benefit from a prior deep reasoning process, instead of making direct judgment of relevance. Recently, remarkable advancements were made in this direction with the introduction of reasoning LLMs, such as OpenAI O1, O3, and DeepSeek R1 \cite{guo2025deepseek}. Particularly, when a complex task is presented, the LLM is prompted to generate long-form thoughts about the problem in the first place. Through this process, the LLM can progressively get close to the correct solution, thus enabling the production of high-quality answers in the end. This operation is conceptualized as the test-time-scaling by recent studies \cite{snell2024scaling}, which has driven major improvements in solving complex problems, such as coding and mathematical proofs. 

With the above inspiration, we propose \textbf{O1 Embedder}, which is designed to introduce a slow-thinking capability for embedding models, akin to that of LLMs. Our approach integrates two essential functions within a single model: \textbf{Thinking} and \textbf{Embedding}. First, it generates useful thoughts towards the input query, which explicitly uncovers the hidden information needs about the query. 
Secondly, it produces a discriminative embedding for the query and the generated thoughts. By incorporating both elements, the resulting embedding enables precise retrieval of relevant documents that are challenging to identify using the query alone.




The training of O1 Embedder is technically challenging given the absence of appropriate long-form thought data for embedding models. To solve this problem, we introduce a \textbf{Data Synthesis} method following an ``\textbf{Exploration}-\textbf{Refinement}'' process. First, we prompt an LLM to explore initial thoughts for a query. Next, we employ a retrieval committee to refine the initial thoughts. Each committee member scores the relevance between initial thoughts and the target document, which indicates their usefulness in retrieving the target document. With the collection of all members' scoring results, the golden thought is selected by majority voting and added to the training set. \textit{As such, we automatically create long-form thoughts of the best retrieval utility for O1 Embedder.}


Building on well-curated long-form thought data, we introduce a \textbf{Multi-Task Training Method}, which fine-tunes a pre-trained LLM into O1 Embedder. Our method introduces two parallel training tasks. One applies supervised fine-tuning for the LLM, which enables the generation of optimal thoughts for an input query. The other one employs contrastive learning, which produces discriminative embeddings to retrieve relevant documents for a thought-augmented query. \textit{With proper optimizations in loss computation and training workflow, these two tasks are seamlessly unified in a cost-efficient manner, leading to effective development of thinking and embedding capabilities throughout the training.} 

The effectiveness of O1 Embedder is comprehensive evaluated. In our experiment, O1 Embedder achieves \textbf{a substantial improvement over existing methods} across a broad range of retrieval tasks, especially those requiring complex reasoning. O1 Embedder also demonstrates \textbf{a strong generalization ability} when applied to out-of-domain scenarios. Finally, O1 Embedder well maintains \textbf{a superior performance across various LLM backbones}, like Llama~\cite{touvron2023llama2openfoundation, grattafiori2024llama3herdmodels}, Mistral~\cite{jiang2023mistral7b}, and Qwen~\cite{grattafiori2024llama3herdmodels}. Our model and code will be made publicly available to advance research in this field. 

To summarize, the main contributions of this paper are highlighted by the following perspectives. 
\begin{itemize}
    \item We propose O1 Embedder, which generates useful thoughts about the input query before producing discriminative embeddings for dense retrieval. To the best of our knowledge, O1 Embedder is the first practice to equip embedding models with thinking capabilities, offering promising insights for future research. 
    \item We design an exploration-refinement approach, producing long-form thoughts with the optimal retrieval utility. We also optimize the multitask training method, which effectively fine-tunes a pre-trained LLM into O1 Embedder. 
    \item We perform comprehensive experiments for a detailed analysis O1 Embedder, whose result verifies the effectiveness and broad applicability of our method. 
\end{itemize}

\begin{figure}[t]
  \centering
  \includegraphics[width=0.80\linewidth]{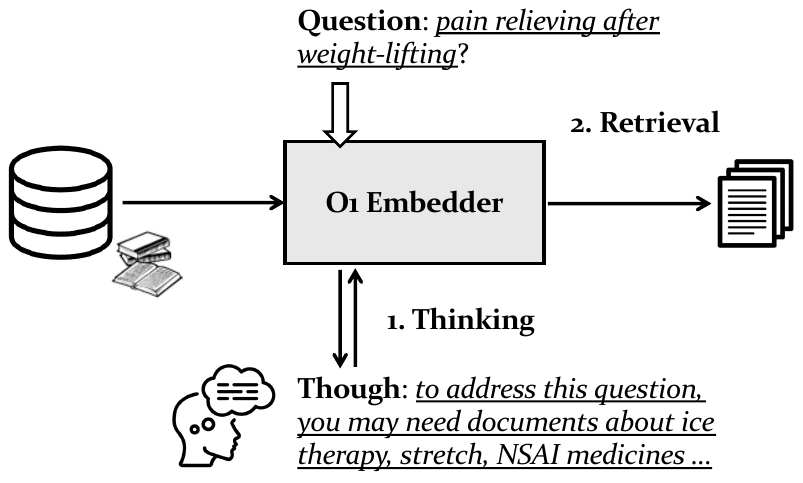}
  \vspace{-5pt}
  \caption{O1 Embedder. First of all, the model generates the thoughts about the question (thinking). Next, the model produces the embedding for dense retrieval (retrieval).}   
  \vspace{-15pt} 
\end{figure}

\section{Related Work} 

\subsection{Dense Retrieval}
Dense retrieval has made significant strides in retrieval precision, driven by the advancements in foundation models and training techniques. Early breakthroughs involved fine-tuning preliminary pre-trained models, such as BERT and RoBERTa \cite{devlin2018bert,liu2019roberta}, for dense retrieval, which already demonstrated competitive performance compared to traditional methods like BM25. At the same time, the scope of dense retrieval was substantially expanded thanks to the adoption of multi-lingual \cite{izacard2022unsuperviseddenseinformationretrieval,chen_bge_2024} and multi-modal pre-trained models \cite{wei2024uniir,zhou2024vista}. The introduction of more advanced training strategies, such as retrieval-oriented adaptation \cite{xiao_retromae_2022,liu_retromae-2_2023,wang_simlm_2023}, hard negative mining \cite{xiong_approximate_2020}, batch size expansion \cite{qu2020rocketqa}, and knowledge distillation from cross-encoders \cite{hofstatter_efficiently_2021}, has continually contributed to the improvement of dense retrieval's performance. 

In addition to the improvement on retrieval accuracy, it becomes increasingly emphasized to develop multi-task retrievers for general-purpose retrieval applications. Recent studies showed that the retrievers' generalization ability can be substantially enhanced by scaling-up the training scale \cite{su2022one} and model architecture \cite{ni_large_2022}. Based on these inspirations, people have made significant expansion of pre-training and fine-tuning tasks, leading to a series of popular retrievers for general-purpose applications, such as BGE, E5, and GTE \cite{wang2022text,li2023towards}. Meanwhile, people also introduce large language models (LLMs) as the retrievers' backbones, which brings forth significant improvements in retrieval performance. For example, RepLLaMA presents a powerful dense retriever by directly fine-tuning a pre-trained Llama \cite{wang2023improving}. Llama2Vec further enhances RepLLaMA by incorporating unsupervised adaptation of the pre-trained Llama \cite{li_llama2vec_2024}. Promptriver~\cite{weller_promptriever_2024}, built on RepLLaMA, equips the retrieval model with the capability to follow instructions. Methods like NV-Embed and ICL-Embedder achieves additional improvement through continual training with extensive fine-tuning data \cite{lee2024nv,li2024making}. Today, LLM-powered retrievers have dominated nearly all major benchmarks in IR-related evaluation. 

Despite these remarkable advancements, existing methods are primarily designed for direct semantic matching in popular applications like web search and question-answering. They still face challenges with zero-shot retrieval in completely new scenarios that differ significantly from their source domains \cite{gao2022precise,zhu2023large}. In addition, they are insufficient for more complex retrieval tasks which require intensive reasoning to identify semantic relationships \cite{su2024bright}. 

\subsection{LLMs' Reasoning Ability} 
The reasoning capabilities of large language models (LLMs) have been significantly enhanced with techniques that simulate human-like problem-solving processes. A major breakthrough in this area is Chain-of-Thought (CoT) \cite{wei2022chain}, which prompts LLMs to tackle complex problems by decomposing them into multiple reasoning steps. Building on this progress, the Self-Consistency method improves reasoning robustness by sampling multiple reasoning paths from the LLM and selecting the final answer through majority voting \cite{feng2024towards}. For scenarios requiring more exploratory reasoning, the Tree of Thoughts (ToT) method \cite{yao2024tree} extends CoT by structuring the problem-solving process as a tree. At each node, the LLM generates candidate intermediate steps, evaluates their feasibility, and backtracks from dead ends. Further advancing this paradigm, Graph of Thoughts (GoT) \cite{besta2024graph} replaces the tree structure with a directed acyclic graph (DAG), enabling LLMs to merge or refine reasoning steps as needed. The reasoning capability of large language models (LLMs), or the "think before action" workflow, represents a new paradigm that sets them apart from traditional language models. In addition to the usual strategies of scaling model size, datasets, and training computation \cite{kaplan2020scaling, hoffmann2022training}, the expansion of inference computation, or test-time scaling \cite{wu2024inference, chen2024llmcallsneedscaling, sardana2023beyond}, becomes another important factor in driving the improvement of LLMs. This capability has been significantly enhanced and showcased by recent reasoning-capable LLMs, such as OpenAI's O1 and O3, DeepSeek's R1 \cite{guo2025deepseek}, and Gemini 2.0\footnote{https://deepmind.google/technologies/gemini/flash-thinking/}. These models adopt a "slow-thinking" approach when handling complex problems: instead of providing an immediate answer, they first generate verbose, structured reasoning before arriving at a final solution. This method has allowed LLMs to achieve elite-level performance in areas like coding and mathematical proofs. 

The reasoning capability also offers a significant advantage in addressing the challenges posed by traditional retrieval methods. However, current embedding models primarily focus on generating discriminative data representations, which leaves the development of reasoning capabilities largely unexplored. 



\section{Method}  
In this section, we first present the problem formulation of O1 Embedder. We then introduce the two main technical contributions of this work: the production of long-form thought data for O1 Embedder, and the multi-task training process of O1 Embedder.

\subsection{Problem Formulation}
Dense retrieval measures the relevance between query and document based on their embedding similarity. Given an query $q$ and document $d$, an embedding model ($\mathcal{M}$) is used to encode them into latent representations $\boldsymbol{v}_q$ and $\boldsymbol{v}_d$: $\boldsymbol{v}_q \leftarrow \mathcal{M}(q)$, $\boldsymbol{v}_d \leftarrow \mathcal{M}(d)$. To retrieve the relevant document $d^*$ from a massive dataset $D$, the following nearest neighbor condition needs to be satisfied: 
\begin{equation}
    d^* = \textsc{argmax}. ~ \{\langle \boldsymbol{v}_q, \boldsymbol{v}_d \rangle | d \in D\}, 
\end{equation}
where $\langle\cdot\rangle$ indicates the inner-product operator. As discussed,  traditional embedding models are insufficient to handle the challenges regarding zero-shot or complex retrieval problems.

\begin{figure*}[t]
  \centering
  \includegraphics[width=0.90\linewidth]{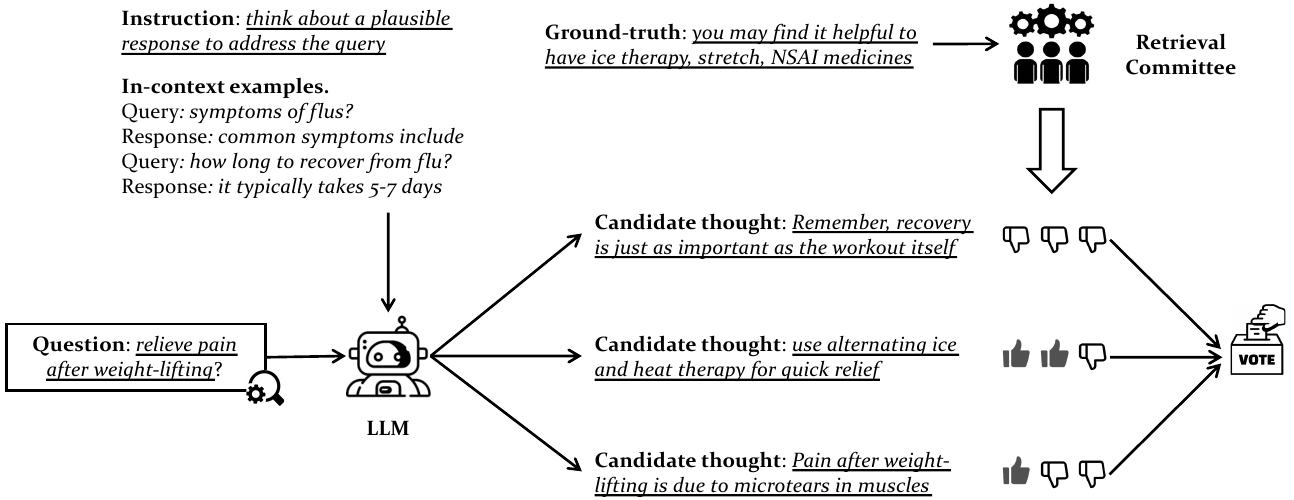} 
  \vspace{-5pt}
  \caption{The production of thought data. In the first step, the LLM is prompted to generate candidates thoughts about the input question based on the instruction and in-context examples. In the second step, the retrieval committee is employed to evaluate the candidates by making comparison with the ground-truth document, i.e. the retrieval target. Finally, the candidate thought receiving the maximum votes is selected and incorporated to the training data.}
  \vspace{-5pt}
  \label{fig:data_produce} 
\end{figure*}

Our method tackles the above challenge by introducing the thinking operation to embedding models. That is to say, the embedding model $\mathcal{M}$ is equipped with two functionalities: thinking $\mathcal{M}.$think($\cdot$) and embedding $\mathcal{M}.$embed($\cdot$). For an input query $q$, the embedding model generates its thoughts ($t$) on how to address the information needs of the query in the first place:  
\begin{equation}
    t_i \leftarrow \mathcal{M}.\text{think}(q), ~ i = 1,...,k
\end{equation}
By revealing critical semantic matching patterns with relevant documents, the generated thoughts are expected facilitate the retrieval process for complex queries. In this place, a total of $k$ thoughts are independently generated with respect to the query, which enables useful patterns to be comprehensively covered. 

On top of the embedding model $\mathcal{M}$ and an aggregation function $\textsc{Agg}$,  the query and its thoughts are jointly transformed into a unified embedding, called the thought-augmented embedding ($\boldsymbol{\hat{v}}_q$): 
\begin{equation}\label{eq:3}
    \boldsymbol{\hat{v}}_q \leftarrow \textsc{Agg}.(q,\{t_i\}_{k};\mathcal{M}.\text{embed})
\end{equation}
As a result, the relevance between query and document is computed with the thought-augmented embedding: $\langle \boldsymbol{\hat{v}}_q, \boldsymbol{v}_d \rangle$. Finally, our problem is formulated as the joint training of thinking and embedding capability of model $\mathcal{M}$, such that the end-to-end retrieval performance is optimized.

\subsection{Data Production} 
\label{section: data production}
The training of O1 Embedder involves two types of data. One is used for the embedding capability, which is made up of queries and their relevant documents, i.e., q-doc tuples. The other one is used for the thinking capability, which includes queries and their thoughts, i.e., q-thought tuples. Unlike q-doc tuples which have been widely existed, there are no available q-thought tuples in reality. To resolve this problem, we propose a data synthesis pipeline, leveraging LLMs' readily equipped reasoning capacity to generate such datasets. Our method follows an ``\textbf{exploration-refinement}'' workflow, as demonstrated in Figure \ref{fig:data_produce}.  

First, we employ a LLM to explore candidate thoughts for a given query $q$. To facilitate proper generations from the LLM, the system prompt is formulated with the following template, where both instruction and examples are incorporated:  
\begin{equation}
    \textsc{Prompt} = \textsc{Task}: \{\mathrm{Ins}\}; ~ \textsc{Examples}: \{\mathrm{E}\}; ~ \textsc{Query}: \{\mathrm{q}\}
\end{equation} 
The instruction is used to explicitly declare the demand for the thinking task; for example, "\textit{think about a plausible response to address the query}". While the examples are introduced to demonstrate the form of desirable output. In this place, we randomly select $m$ samples from the training set of q-doc dataset: 
\begin{equation}
    \mathrm{E} = \{ \textsc{Query}: q_i, ~ \textsc{Response}: d_i \}_m. 
\end{equation} 
Note that although the relevant document $d$ for query $q$ may seem like a trivial solution, it is unsuitable to serve as a thought. This is because the generated thought will be further used in embedding task, whose goal is to discriminate the relevant document $d$ based on the thought-augmented embedding. If $d$ (or any rephrased version of it) is used, the training process would be circumvented, ultimately leading to the collapse of the embedding task. 

The generated thoughts may not always enhance retrieval performance due to potential hallucinations by the LLM. To ensure the inclusion of useful thoughts, the exploration process is repeated multiple times, generating several independent thoughts for the given input query. To identify the most useful thoughts, a quality-control mechanism is introduced to filter the generated candidates. Specifically, we employ a diverse set of retrievers, denoted as $R$. For each retriever $r \in R$, a similarity score is computed between a relevant document $d$ and a thought $t_i$: $\sigma^r(t_i, d)$. The thought with the highest similarity score is selected by each retriever, i.e., $t^*_r \leftarrow \textsc{argmax}(\{\sigma^r(t_i, d)\}_{i=1...k})$. Finally, a majority voting process is conducted to determine the most useful thought. The thought that receives the highest number of nominations from the retrievers is selected as the final result: $t \leftarrow \textsc{Voting}\{t^*_r\}_{r \in R}$. 

By applying the above data synthesis workflow to an existing q-doc dataset: $D = \{(q_i,d_i)\}_N$, we can obtain an thought-augmented dataset composed of \textbf{q-thought-doc triplets}: $\hat{D} = \{(q_i, t_i, d_i)\}_N$, which offers long-form thoughts of the optimal retrieval utility. 


\subsection{Multi-Task Training} 
The O1 embedder is built upon a pre-trained LLM, leveraging the model’s inherent generation and reasoning abilities. Additionally, LLMs also show strong potential for fine-tuning as discriminative embedding models \cite{luo2024large, zhu_large_2024}. We apply the following multitask training for a pre-trained LLM backbone, which establishes its thinking capability via supervised behavior cloning and embedding ability through contrastive learning. 


\subsubsection{Behavior cloning}  
The foundation model is trained to generate thoughts for the input query through supervised fine-tuning. Given the dataset of q-thought-doc triplets: $\hat{D} = \{(q_i, t_i, d_i)\}_N$, a training sample $x_i$ is formulated with the template below: 
\begin{equation}
    x_i =\text{<query>} q_i \text{<thought>} t_i  \text{</s>},  
\end{equation} 
where <query>, <thought>, </s> are the special tokens to landmark the query, thought, and the completion of generation, respectively. 

With the formulation of above training samples, the model is fine-tuned w.r.t. the following \textbf{generation loss}: 
\begin{equation}
    \mathcal{L}_{gen} = - \sum_{x_i} \sum\nolimits_{j=|q_i|}^{|q_i|+|t_i|} \log P(x_{i,j} | x_{i,<j}),
\end{equation}
where the next-token-prediction loss is minimized for each of the tokens starting from the beginning of the thought (i.e. $j \geq |q_i|$).  

\begin{figure*}[t]
  \centering
  \includegraphics[width=0.95\linewidth]{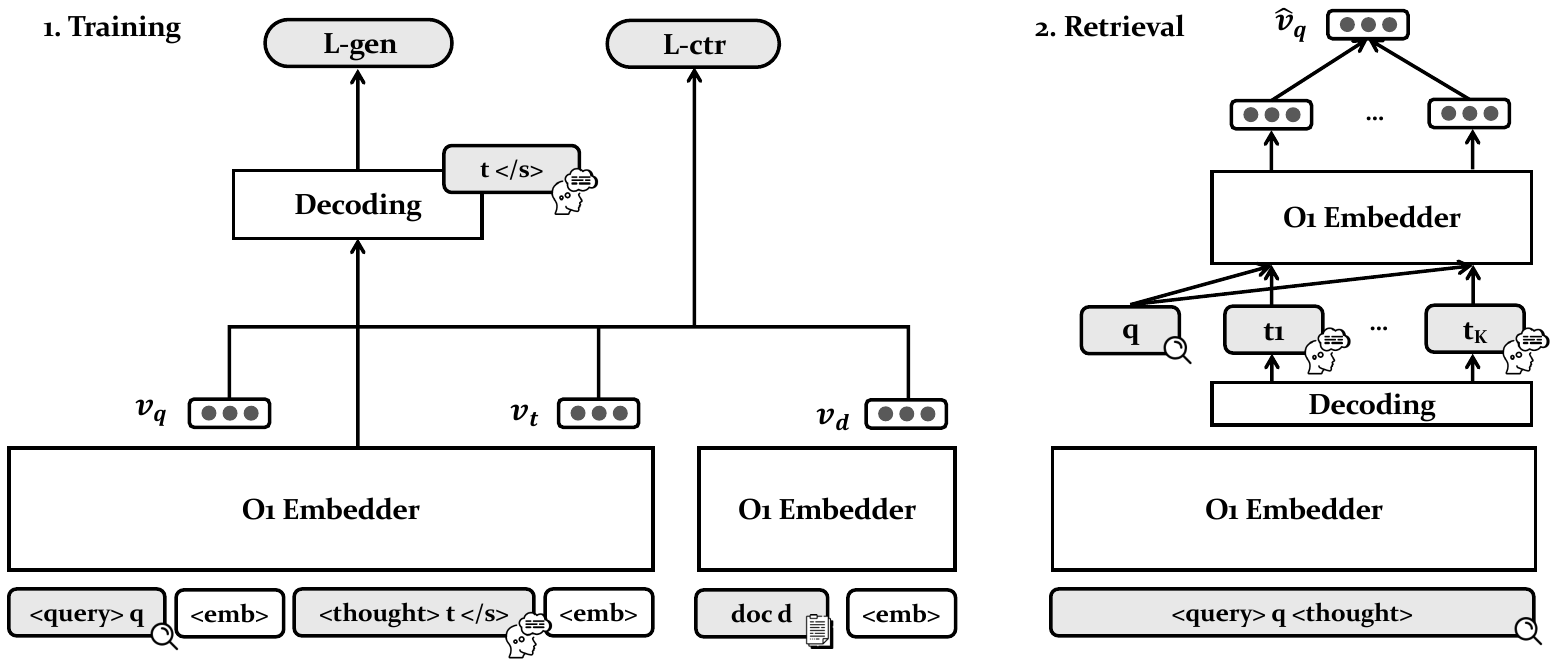}
  \vspace{-5pt}
  \caption{Training and Retrieval process of O1 Embedder. During the training process, O1 embedder minimizes two losses: the generation loss while decoding the thought, and the contrastive loss while discriminating the target document. During the retrieval process, multiple thoughts are generated for the query. The thoughts are used to produce thought-augmented queries, which are independently encoded by O1 Embedder and aggregated for retrieval.} 
  \label{fig:train_retrieve}
\end{figure*} 

\subsubsection{Contrastive learning} 
The pre-trained LLM is also fine-tuned to distinguish relevant documents from a query based on its generated embeddings. Traditionally, LLM-based embedders utilize the </s> token for text embedding. However, the </s> token has been designated to indicate the completion of generation process in our approach. As a result, incorporating an extra embedding task could lead to the collapse of training process. {To avoid mutual interference between the two parallel training tasks, we employ another \textbf{special token <emb>} and append it to the end of input $x$ (following </s>) to compute the text embedding:} 
\begin{equation}
    \boldsymbol{v}_x \leftarrow \text{LLM}(x;\text{<emb>})[-1]
\end{equation} 
\textit{This simple modification substantially increases the compatibility, which is crucial to maintain the successful running of joint training process.} Considering that people may want to leverage thought-augmented embedding to handle complex retrieval tasks while still relying on basic query embedding to process simple retrieval tasks, we generate the two forms of query embeddings simultaneously to identify the relevant document $d_i$. As a result, we perform the following composite contrastive learning, where two \textbf{contrastive losses} are calculated based on the query and the thought-augmented query of each training sample, respectively:  
\begin{equation}
    \mathcal{L}_{ctr} = - \sum_{i} \frac{\exp(\boldsymbol{v}_{q_i}^T \boldsymbol{v}_{d_i})}{\sum_{d'}\exp( \boldsymbol{v}_{q_i}^T \boldsymbol{v}_{d'})} + \frac{\exp(\boldsymbol{v}_{\hat{q}_i}^T \boldsymbol{v}_{d_i})}{\sum_{D'}\exp( \boldsymbol{v}_{\hat{q}_i}^T \boldsymbol{v}_{d'})}
\end{equation}
In this place, $\hat{q}_i$ indicates the thought-augmented query: $\hat{q}_i \leftarrow q_i + t_i$, $D'$ stands for the collection of negative samples, including both in-batch negative samples and hard negative samples introduced by a pre-trained embedder.  

\subsubsection{Joint training}
The model is trained to minimize the linear combination of generation loss and contrastive loss: 
\begin{equation}
    \mathcal{L} = \lambda \mathcal{L}_{gen} + (1-\lambda) \mathcal{L}_{ctr}
\end{equation} 
To enable precise retrieval in downstream scenarios, contrastive learning must be conducted with a large training batch. However, the native parallel running of two training tasks requires significant GPU memory, which severely limits the achievable batch size. To address this challenge, we propose a \textbf{memory-efficient joint training} to handle both tasks (Figure \ref{fig:train_retrieve}). Specifically, for each training sample $x_i = (q_i, t_i, d_i)$, we encode it once and share the encoding result between the two tasks. The generative loss is calculated based on each of the output embeddings from $t_i$ tokens; while the contrastive loss is derived based on the output embeddings from <emb> tokens. This allows the generation task to consume almost no additional memory when trained alongside contrastive learning task, thereby enabling a substantial increase in batch size.  

\subsection{Retrieval} 
The well-trained O1 embedder $\mathcal{M}$ is applied for retrieval tasks through thinking and embedding. First, O1 embedder is prompted to generated multiple thoughts towards the input query which comprehensively uncover the query's hidden information needs: $t_i \leftarrow \mathcal{M}.\text{think(q)}, i = 1, ...  , k$. Next, the thought-augmented queries are independently encoded and aggregated. In this place, we simply adopt mean pooling as the aggregation function in Eq. \ref{eq:3}, which produces the following thought-augmented embedding:  
\begin{equation}
    \boldsymbol{\hat{v}}_q \leftarrow \sum_{k} \mathcal{M}.\text{ enc($q,t_i$)} \Bigg/  {k} 
\end{equation} 
Finally, the top-N documents $D^*$ are retrieved based on their embedding similarity with $\boldsymbol{\hat{v}}_q$: $D^* \leftarrow \text{top-N}(\{\langle \boldsymbol{\hat{v}}_q,\boldsymbol{v}_d \rangle|D\})$.



\section{Experiment} 
We make comprehensive evaluation of our approach with a focus on the following research questions. 
\begin{itemize}
  \item[\textbf{RQ 1.}] Whether O1 Embedder can outperform popular baseline retrievers after fine-tuning? 
  \item[\textbf{RQ 2.}] Whether O1 Embedder can be effectively generalized to out-of-domain scenarios? 
  \item[\textbf{RQ 3.}] How much does the thinking operation help? 
\end{itemize}
With these research questions, we design our experimental studies, whose settings are presented as follows. 

\subsection{Settings}

\begin{table*}[ht]
\begin{tabular}{c|c|c|cc|c|c}
\hline
                            & \multirow{2}{*}{method} & \multirow{2}{*}{model size} & \multicolumn{2}{c|}{MS MARCO Dev} & DL'19         & DL'20         \\ \cline{4-7} 
                            &                         &                             & MRR@10          & Recall@1k       & nDCG@10       & nDCG@10       \\ \hline
Sparse                      & BM25\cite{thakur_beir_2021}                    & -                           & 18.4            & 85.3            & 50.6          & 48.0          \\ \hline
\multirow{5}{*}{BERT-based} & ANCE\cite{xiong_approximate_2020}                    & 125M                        & 33.0            & 95.9            & 64.8          & 61.5          \\
                            & TAS-B\cite{hofstatter_efficiently_2021}                   & 55M                         & 34.3            & 97.6            & 72.2          & 69.2          \\
                            & coCondenser\cite{gao_unsupervised_2022}             & 110M                        & 38.2            & 98.4            & 69.8          & 68.4          \\
                            & SimLM\cite{wang_simlm_2023}                   & 110M                        & 41.1            & 98.7            & 71.4          & 69.7          \\
                            & RetroMAE\cite{liu_retromae-2_2023}                & 110M                        & 41.6            & 98.8            & 68.1          & 70.6          \\ \hline
\multirow{4}{*}{LLM-based}  & RepLLaMA\cite{ma_fine-tuning_2024}                & 7B                          & 41.2            & 99.4            & 74.3          & 72.1          \\
                            & Promptriever\cite{weller_promptriever_2024}            & 7B                          & 41.0            & 99.4            & 73.2          & 72.3          \\
                            & \textbf{O1 embedder w/o T}  & 7B                          & 41.7            & 99.4            & 73.7          & 72.3          \\
                            & \textbf{O1 embedder}               & 7B                          & \textbf{43.1}   & \textbf{99.5}   & \textbf{75.3} & \textbf{74.4} \\ \hline
\end{tabular} 
\vspace{5pt} 
\caption{In-domain evaluation results evaluated on MS MARCO, DL'19, and DL'20. O1 embedder w/o T indicates the variational form of O1 embedder, which disables the thinking operation but uses the same embedding model.} 
  \label{tab:MSMARCO}
\end{table*}

\subsubsection{Datasets}
O1 embedder is trained by the thought-augmented queries created from the MS MARCO (passage retrieval) dataset \cite{bajaj2018msmarcohumangenerated}. The well-trained model is evaluated based on both in-domain and out-of-domain datasets. For in-domain evaluation, we utilize MS MARCO (dev), TREC DL19~\cite{craswell2020overviewtrec2019deep}, and TREC DL20~\cite{craswell2021overviewtrec2020deep} datasets. For out-of-domain evaluation, we incorporate the following eight question-answering datasets from BEIR~\cite{thakur_beir_2021}, including SciFact~\cite{wadden-etal-2020-fact}, TREC-COVID~\cite{10.1145/3451964.3451965}, DBpedia~\cite{10.1145/3077136.3080751}, NQ~\cite{kwiatkowski-etal-2019-natural}, HotPotQA~\cite{yang2018hotpotqa}, FiQA~\cite{maia201818}, Touche~\cite{bondarenko2020overview}, FEVER~\cite{thorne-etal-2018-fever}, along with a popular dataset on code-search: CosQA~\cite{huang-etal-2021-cosqa}. All of these datasets consist of asymmetric retrieval tasks, where the query and document are presented in very different forms. As a result, the relationships between query and document can be more effectively identified through appropriate reasoning. We exclude common paraphrasing datasets, such as Quora, as they only involve simple similarity comparisons. Following prior works \cite{ma_fine-tuning_2024, weller_promptriever_2024}, we use MRR@10 and Recall@1k as metrics for MS MARCO-related tasks, and NDCG@10 for other datasets. Evaluations on o.o.d. datasets strictly adhere to the BEIR protocol, which prohibits task-specific fine-tuning.

\subsubsection{Baselines} 
We choose a wide variety of popular retrievers as our baselines, such as BM25, a commonly used sparse retrieval method, and ANCE \cite{xiong_approximate_2020}, TAS-B \cite{hofstatter_efficiently_2021}, coCondenser \cite{gao_unsupervised_2022}, SimLM \cite{wang_simlm_2023}, which fine-tune BERT-based pre-trained models using MS MARCO dataset. We also introduce the LLM-based methods, including RepLLaMA \cite{wang_text_2024}, Promptriver \cite{weller_promptriever_2024}. RepLLaMA fine-tunes a pre-trained LLM on MS MARCO, resulting in superior retrieval performance across various downstream tasks. While Promptriever builds on RepLLaMA by enhancing the model’s instruction-following capabilities, which further improves the retrieval performance on top of tailored prompts. Both LLM-based methods are fine-tuned from the same pre-trained backbone (Llama-2-7B) as our default setting, thus ensuring a fair comparison in terms of model scale. Note that we exclude several other popular retrievers from our evaluation, including BGE~\cite{xiao_c-pack_2024}, E5~\cite{wang2022text}, M3~\cite{chen_bge_2024}, and recent LLM-based methods like E5-Mistral~\cite{wang2023improving}, ICL-Embedder~\cite{li_making_2024}, and NV-Embed~\cite{lee2024nv}. These models utilize significantly more training data beyond MS MARCO (the only training dataset used by our method and our included baselines), many having strong overlaps with the evaluation tasks on BEIR. This overlap makes it difficult to assess zero-shot retrieval performance in out-of-distribution (o.o.d.) settings. Additionally, these methods rely on different training datasets, which makes it impossible to maintain a fair comparison. 


\subsubsection{Implementation Details}
During the data preparation stage, we leverage a powerful open-source LLM: Llama-3.1-70B-Instruct\footnote{https://huggingface.co/meta-llama/Llama-3.1-70B-Instruct}, to generated the candidate thoughts. We employ BM25, BGE-EN-large-v1.5, GTE-large, and Stella-EN-1.5B-v5, to serve the retrieval committee. The evaluation is primary made based on a Llama-2-7B backbone \cite{touvron2023llama2openfoundation}, with other alternative LLMs analyzed in the extended study. The training process follows RepLLaMA's recipe \cite{ma_fine-tuning_2024}, where all projection layers (q\_proj
k\_proj v\_proj o\_proj gate\_proj down\_proj up\_proj) of the LLM are fine-tuned via LoRA, with rank set to 32 and alpha set to 64. We used BF16 for training, with learning rate set to \( 1 \times 10^{-4} \). The training process takes place on 8xA800 GPUs, with a batch size set to 64 (8 per-device). We introduce 15 hard negatives for each query. The maximum query length was set to 32, and the maximum passage length was set to 192. The max tokens were set to 256 for thought generation.

\subsection{Main Results}
\subsubsection{In-domain Performance}
\label{subsection: Superviesd Performance} 
The in-domain evaluation results are presented in Table \ref{tab:MSMARCO}, where the O1 Embedder outperforms all other models across all evaluation metrics, including MRR@10, Recall@1k, and nDCG@10, on the MSMARCO, DL'19, and DL'20 datasets. Specifically, our model achieves an MRR@10 of 43.1 (\textbf{+1.9\%} improvement) and a Recall@1k of 99.5 on MSMARCO, along with nDCG@10 scores of 75.3 (\textbf{+1.0\%} improvement) and 74.4 (\textbf{2.1\%} improvement) on DL'19 and DL'20, respectively, compared to RepLLaMA. These results demonstrate the effectiveness of our model and its enhanced reasoning capabilities in retrieval tasks. 
Furthermore, when comparing model sizes, it is clear that larger models, such as the LLM-based RepLLaMA and Promptriever, generally outperform smaller BERT-based models. For example, RepLLaMA achieves an nDCG@10 of 74.3 on DL'19 and 72.1 on DL'20, surpassing all smaller models. The O1 Embedder without thinking (O1 embedder w/o T), a variation that disables the reasoning operation, yields similar results as RepLLaMA, suggesting that the embedding capability is effectively established through multi-task training. However, with the addition of thinking, the full O1 Embedder demonstrates a substantial improvement, highlighting the power of the "thinking" enhancement.

\subsubsection{O.O.D. Performance} 
\label{subsection: Zero-shot Generalization}

\begin{table*}[]

\begin{tabular}{c|c|ccccccccc|c}
\hline
Method          & Model size & T-Covid & NQ   & HQA & FiQA & Touche & DBPedia & FEVER & SciFact & CosQA & Average \\ \hline
Contriever      & 0.1B       & 59.6       & 49.8 & 63.8     & 32.9 & 23.0   & 41.3    & 75.8  & 67.7    & 14.2  & 47.6    \\
CPT-L           & 6B         & 56.2       & -    & 64.8     & 45.2 & 30.9   & 41.2    & 75.6  & 74.4    & -     & -       \\
CPT-XL          & 175B       & 64.9       & -    & 68.8     & \textbf{51.2} & 29.1   & 43.2    & 77.5  & 75.4    & -     & -       \\
OpenAI-Ada-002  & -          & 81.3       & 48.2 & 65.4     & 41.1 & 28.0   & 40.2    & 77.3  & 73.6    & 28.9  & 53.7    \\
RepLLaMA        & 7B         & 84.7       & 62.4 & 68.5     & 45.8 & 30.5   & 43.7    & 83.4  & 75.6    & 32.3  & 58.5    \\
Promptriver     & 7B         & 84.6       & 62.6 & 69.5     & 46.6 & 32.0   & 45.2    & 82.8  & 76.3    & 32.8  & 59.2    \\ \hline
\textbf{O1 embedder w/o T} & 7B         & 84.5       & 62.9 & 69.8     & 45.0 & 33.8   & 44.4    & 82.5  & 75.8    & 32.9  & 59.1    \\
\textbf{O1 embeder}    & 7B         & \textbf{85.6}       & \textbf{66.8} & \textbf{72.8}     & 46.6 & \textbf{36.7}   & \textbf{47.3}    & \textbf{84.9}  & \textbf{77.4}    & \textbf{34.1}  & \textbf{61.4}    \\ \hline
\end{tabular}
\vspace{5pt}
\caption{Out-of-domain evaluation results measured by nDCG@10.}
  \label{tab:BEIR}
\end{table*} 

The out-of-distribution (o.o.d.) evaluation results are shown in Table \ref{tab:BEIR}, where the O1 Embedder consistently outperforms across all nine datasets. On average, our model achieves a \textbf{2.3\%} improvement over the baseline, which is a significant boost and underscores the strong generalization capabilities of our approach. Additionally, for each of the nine datasets, our model sets the highest performance except for FiQA, where it falls behind CPT-XL 175B. This highlights that our model excels across various scenarios, especially the retrieval tasks involving complex reasoning, such as HotPotQA (multi-hop question-answering) and CosQA (code-search).

Several interesting conclusions can be derived from the table. The thinking mechanism leads to more significant improvements on certain OpenQA datasets. For instance, the NQ dataset shows a \textbf{3.9\%} improvement, while HotPotQA sees a \textbf{3.0\%} boost. However, while there are improvements in some specialized domains, they are not as substantial. For example, in TREC-Covid (medical domain), FiQA (financial domain), and SciFact (scientific domain), the improvements brought by the thinking mechanism are only 0.9\%, 1.7\%, and 1.6\%, respectively. We believe this is because LLMs perform well on general OpenQA questions after training on large-scale corpora, but often lack specialized domain data. As a result, the generated content can sometimes be noisy or even hallucinated. While the thinking mechanism still provides some enhancement, its impact is not as pronounced in these domains. Thus, refining the generated thoughts will be an important issue for future research. 

\subsubsection{Impact of Thought} 


The comparisons between the O1 Embedder and the O1 Embedder w/o T in Tables~\ref{tab:MSMARCO} and~\ref{tab:BEIR} clearly demonstrate the significant impact of incorporating the thinking mechanism. Specifically, on the MS MARCO benchmark (Table~\ref{tab:MSMARCO}), the O1 Embedder shows substantial improvements across all metrics, with notable gains in MRR@10 (+1.4\%) and nDCG@10 for both DL'19 (+1.6\%) and DL'20 (+2.1\%). Similarly, in the zero-shot evaluation (Table~\ref{tab:BEIR}), the O1 Embedder outperforms its counterpart across all datasets, achieving an average nDCG@10 score of 61.4—2.3 points higher than the O1 Embedder w/o T. This performance boost can be attributed to the thinking mechanism, which allows the model to generate additional contextual information during the encoding process. By incorporating this supplementary information, the model enhances its understanding of the user’s query, leading to more accurate and effective retrieval. 
Additionally, it is worth noting that even in its "fast" mode (represented by O1 Embedder w/o T), the model achieves results comparable to those of RepLLaMA. This highlights that our model maintains competitive retrieval performance, even without the generated reasoning, underscoring its flexibility and adaptability.

\subsubsection{Summary for main result} 
Based on the above discussion, we come to the following conclusions in response to RQ 1-3:  

\begin{itemize}
  \item[\textbf{Con 1.}] Our model significantly outperforms both BERT-based and LLM-based models for in-domain evaluations, indicating that the model's thinking and embedding capabilities are effectively learned through the multi-task training process.  
  \item[\textbf{Con 2.}] Our model well maintains superior performances throughout different o.o.d. tasks, which verifies our strong generalization ability. This advantage is especially pronounced when handling those challenging problems. 
  \item[\textbf{Con 3.}] The thinking operation brings forth significant help, as it substantially improves upon the retrieval performance from the vanilla embedding model. 
\end{itemize}

\subsection{Extended Analysis} 
With the verification of O1 Embedder's overall effectiveness in the previous discussions, we perform extended studies in this section, where the following detailed problems are analyzed: 


\begin{itemize}
  \item[\textbf{RQ 4.}] How much does our joint multi-task training strategy contribute to the model’s performance? 
  \item[\textbf{RQ 5.}] Whether our method stays robust to different model architectures and parameter settings? 
  \item[\textbf{RQ 6.}] Why does our method achieves such significant improvements in its retrieval performance? 
\end{itemize} 

The corresponding issues are discussed in subsections \ref{subsection: RepLLaMA with thinking}, \ref{subsection: Robustness}, and \ref{subsection: Case Study}, respectively. 


\subsubsection{Joint training}
\label{subsection: RepLLaMA with thinking}

\begin{table}[ht]

\begin{tabular}{c|ccc|ccc}
\hline
           & \multicolumn{3}{c|}{RepLLaMA} & \multicolumn{3}{c}{O1 embedder} \\ \cline{2-7} 
           & base    & with T    & $\Delta$   & base  & with T  & $\Delta$  \\ \hline
DL'19       & 74.3  &   72.4    &   -1.9    &   73.7    &   75.3    &   +1.6    \\
DL'20       & 72.1  &   72.6    &   +0.5    &   72.3    &   74.4    &   +2.1    \\  \hline
Trec-Covid & 84.7    & 79.7      & -5.0      & 84.5  & 85.6    & +1.1    \\
NQ         & 62.4    & 65.1      & +2.7     & 62.9  & 66.8    & +3.9    \\
HotPotQA   & 68.5    & 71.7      & +3.2     & 69.8  & 72.8    & +3.0      \\
FiQA       & 45.8    & 40.2      & -5.6    & 45.0    & 46.6    & +1.6    \\
Touche     & 30.5    & 33.3      & +2.8     & 33.8  & 36.7    & +2.9    \\
DBPedia    & 43.7    & 44.2      & +0.5     & 44.4  & 47.3    & +2.9    \\
FEVER      & 83.4    & 85.5      & +2.1     & 82.5  & 85.0      & +2.5    \\
SciFact    & 75.6    & 74.2      & -1.4    & 75.8  & 77.4    & +1.6    \\ 
CosQA      & 32.3    & 33.0     & +0.7      & 32.9  & 34.1  & +1.2      \\ \hline
Avg        & 61.2    & 61.1      & -0.1    & 61.6  & 63.8    & +2.2    \\ \hline
\end{tabular}
\vspace{5pt}
\caption{Exploration of joint training. With the joint training of thinking and embedding ability, O1 embedder achieves a substantial improvement in retrieval performance. In contrast, RepLLaMA directly leverages the thoughts generated by GPT-4o-mini (using the same prompt as O1 embedder), which leads to a suboptimal performance due to the incompatibility of the two modules. In this table, "Base" denotes retrieval directly with query, "with T" denotes retrieval using the query with thought, "$\Delta$" denotes the improvement.}
\vspace{-15pt}
\label{tab: RepLLaMA+}
\end{table}

To analyze the impact from joint training, we make analysis on whether existing retrieval models can make effective use of the generated thoughts in a training-free manner. For this purpose, we introduce a stand-alone generator for a well-trained retriever, which generates thoughts for its presented queries. In our experiment, we leverage RepLLaMA as the retriever and GPT-4o-mini as the generator for our experiments. 

The results of this approach are shown in Table \ref{tab: RepLLaMA+}. While incorporating thoughts results in mild improvements on datasets like NQ and HotPotQA, it causes declines on others, such as Trec-Covid and FiQA. Overall, this approach leads to only minor gains on some tasks but results in a slight decrease of 0.1 in overall performance. In contrast, our model consistently improves the retrieval performance across all datasets. This suggests that, with joint training, our model can better utilize the generated thoughts. The untrained RepLLaMA model, however, appears to be negatively impacted by the potential noise within the generated thoughts, leading to worse results, particularly in specialized domains like Trec-Covid, FiQA, and SciFact. In brief, the above observation indicates that our method not only generates useful thoughts for retrieval, but also learns to make effective use of the thoughts through joint training. 


    

\subsubsection{Robustness}
\label{subsection: Robustness}

In this section, we verify the robustness of our method from two perspectives. First, we implement our method based on different pre-trained architectures, including Llama, Mistral, Qwen. Second, we also introduce LLMs of different sizes (ranging from 0.5B to 8B) for evaluation. 

\begin{table}[ht]

\begin{tabular}{c|ccc|c}
\hline
               & \multicolumn{3}{c|}{in-domain} & o.o.d. \\
               & MS MARCO       & DL'19  & DL'20  &  AVG  \\ \hline
Llama-2-7B     & 43.1          & 75.3   & 74.4   & 61.4      \\
Mistralv0.3-7B & 43.5          & 77.0   & 75.6   & 61.4      \\
Llama-3.1-8B   & 43.5          & 76.2   & 74.5   & 61.6      \\
Qwen2.5-7B     & 43.3          & 76.4   & 74.7   & 61.2      \\
Qwen2.5-3B     & 42.5          & 76.3   & 74.5   & 60.3      \\
Qwen2.5-1.5B   & 41.9          & 74.0   & 73.5   & 58.7      \\
Qwen2.5-0.5B   & 40.5          & 73.6   & 71.4   & 55.4       \\ \hline
\end{tabular}
\vspace{5pt}
\caption{The impact from using different backbone models of variant pre-trained architectures and model sizes. The detailed scores for the zero-shot evaluation are presented in Appendix~\ref{app:detail score}. O1 Embedder well maintains a strong performance throughout these settings.}
\vspace{-15pt}
\label{tab: base model}
\end{table}


\textbf{Impact of different model backbone}. The previous experiments primarily used Llama-2-7B as the backbone, which is consistent with RepLLaMA and promptriever. To verify the generalizability of our approach, we repeat the same experiment with implementations on different backbone LLMs. 
The results in Table~\ref{tab: base model} demonstrate our effectiveness across different settings. Notably, our approach well maintains a strong retrieval performance in both in-domain and out-of-domain evaluations. This observation suggests that our method is generally effective with various architectures, regardless of their difference in pre-trained capabilities.  


\textbf{Impact of different model sizes}. We can also clearly observe the significant impact of model size on performance in Table~\ref{tab: base model}. As the size of the Qwen2.5 models decreases, there is a noticeable drop in effectiveness across all evaluated datasets. While the 3B model performs similarly to the 7B model, further reductions in size lead to more pronounced performance declines. This is partly because larger models tend to have better generalization capabilities, benefiting from training on more extensive corpora during pre-training.
It's worth noting that even a 1.5B model, through retrieval with thought, performs comparably to the 7B RepLLaMA, which further highlights the advantage of our approach.

\subsubsection{Case Study}
\label{subsection: Case Study}

\begin{table}[ht]
\begin{tabular}{p{0.9\linewidth}}
\hline
\textbf{Query:}   what age was martin luther king when he was admitted      \\
\hline
\textbf{Thought:}   Martin Luther King Jr. was admitted to \textcolor{softgreen}{\textbf{Morehouse College in Atlanta}}, Georgia \textcolor{softgreen}{\textbf{at the age of 15}}. He attended the college from \textcolor{softgreen}{\textbf{1944}} to 1948, where he earned a Bachelor of Arts degree in sociology.   \\
\hline
\textbf{Positive Doc:}   Dr. Martin L. King, Jr. and His Mentors: A Message for America Today If it was not for Benjamin Mays... Benjamin Mays was the president of \textcolor{softgreen}{\textbf{Morehouse College in Atlanta}} when he met Martin Luther King, Jr. In \textcolor{softgreen}{\textbf{1944}}, Martin Luther King was admitted to the college \textcolor{softgreen}{\textbf{at age 15}}. ...   \\
\hline
\end{tabular}
\vspace{5pt}
\caption{Examples of the original query, thinking content and the positive document. The similar patterns between gnenrated content and the groundtruth are marked in \textcolor{softgreen}{\textbf{green}}.}
\label{tab:case1}
\end{table}

In Table~\ref{tab:case1}, we demonstrate an example of the generated thought and the ground-truth document to a complex multi-hop query. In this case, the query asks about Martin Luther King's age upon his admission to college. Our thought generated effectively uncovers useful contextual information, highlighting that King was admitted to Morehouse College at the age of 15 in 1944. This generated content not only answers the query directly but also enriches the context by providing additional details about the college and the timeframe of his attendance.  \textbf{By generating such useful patterns, the embedding model can obtain crucial information related to the query, which results in a more precise retrieval result}. We include more case analysis for O1 Embedder in Appendix~\ref{section:additional_case_study} and ~\ref{app: complex}. 




\section{Conclusion} 
In this paper, we introduce O1 Embedder, a novel retrieval model that performs slow-thinking before executing retrieval actions. This approach allows the model to better understand the underlying information needs within the query, aiding in the identification of semantic relevance for complex retrieval tasks. Leveraging our tailored data production method, we generate long-form thoughts optimized for retrieval utility. Additionally, our proposed multi-task training method effectively establishes both the model's thinking and embedding capabilities. We conduct comprehensive experiments on popular evaluation benchmarks, and the results demonstrate that O1 Embedder significantly outperforms existing methods, achieving substantial improvements in retrieval performance across both in-domain and out-of-domain scenarios. 
Our work lays the foundation for future research in advanced retrieval models with reasoning capabilities. Future directions include expanding the reasoning process to multi-round interactions, exploring lightweight distillation techniques, and applying the approach to other retrieval tasks. The O1 Embedder marks a promising paradigm for next-generation IR systems, showcasing the potential of integrating the classic dense retrieval methods with large language models' outstanding reasoning abilities. 
\clearpage 

\bibliographystyle{ACM-Reference-Format}
\bibliography{ref}

\clearpage

\appendix




\section{Prompts}
\label{section:prompts}

\begin{table}[ht]
\begin{tabular}{p{0.9\linewidth}}
\hline
System Prompt \\
\hline
   You are a helpful assistant. Your anwer should be follow the task description. Do not ask the user for further clarification. Don't repeat the query, just give the response.   \\
\hline
\hline
User Prompt \\
\hline
Task: \newline Think about a plausible response to address the query \newline \newline Examples: \newline    Query: love ranch movie cast    \newline
Response: Love Ranch Love Ranch. Love Ranch is ...<omit> \newline
Query: hanover va real estate tax   \newline
Response: Hanover County Virginia Property Tax Median Income In. Hanover County, Virginia. The median property tax (also known as real estate tax) in Hanover County is \$1,922.00 per year, based on ...<omit> \newline
Query: what is polymyositis symptoms    \newline
Response: - Polymyositis is a systemic disease, which means it affects the whole body. Muscle weakness and tenderness can ...<omit> 
\newline \newline   Query: \newline what does physical medicine do \newline    Response:   \\
\hline
\hline
Response \\
\hline
Physical medicine, also known as physiatry, is a branch of medicine that focuses on helping patients with physical disabilities or impairments to achieve maximum function and independence. Physiatrists, or physical medicine doctors, use a multidisciplinary approach to diagnose and treat a wide range of conditions, 
including musculoskeletal disorders, neurological conditions, and chronic pain.Physical medicine doctors use various techniques, such as physical therapy, occupational therapy, and pain management, to help patients recover from injuries or illnesses, manage chronic conditions, and improve their overall quality of life. They may also use advanced technologies, such as electromyography (EMG) and nerve conduction studies, to diagnose and treat conditions.Some common conditions treated by physical medicine doctors include back pain, arthritis, spinal cord injuries, stroke, and traumatic brain injuries. Physical medicine doctors work with patients to develop personalized treatment plans that address their unique needs and goals, with the aim of restoring function, reducing pain, and improving overall well-being.   \\  
\hline
\end{tabular}
\caption{Example of the Prompt and the Response of the teacher model for data generation. For presentation purposes, we use <omit> to omit the long documentation in the example.}
\label{tab:prompt}
\end{table}

Table \ref{tab:prompt} shows the example of complete prompts and the response of the teacher model used to generate the training data. For each specific query, we randomly select 3 query and positive document pair to form the in-context examples. During the data generation process, we found that a significant portion of the responses duplicated the query provided. So we add "\textit{Don’t repeat the query, just give the response.}" in the system prompts.
\section{Detailed Scores of Different Base Models}
\label{app:detail score}
Table~\ref{tab:detailed scores} provides the detailed scores of different base model for zero-shot evaluation in section~\ref{subsection: Robustness}. We can see that although the results of different base models fluctuate a bit in each dataset, there is not much difference in the overall performance of models of the same scale. While when the model parameters decreased, there was a significant decrease in the final results. This suggests that larger models are more capable of generating and exploiting high-quality thoughts and have greater generalization.

\begin{table*}[ht]
\scalebox{1.0}{

\begin{tabular}{c|ccccccccl|c}
\hline
                            & Trec-Covid & NQ   & HotPotQA & FiQA & Touche & DBPedia & FEVER & SciFact & CosQA & Average \\ \hline
O1 Embedder(Llama3.1-8B)    & 83.9       & 67.4 & 74.4     & 47.4 & 35.0   & 47.2    & 86.5  & 77.2    & 35.6     & 61.6    \\
O1 Embedder(Mistralv0.3-7B) & 85.9       & 67.4 & 72.6     & 46.1 & 36.7   & 46.9    & 84.8  & 76.2    & 36.0     & 61.4    \\
O1 Embedder(Llama2-7B) & 85.6       & 66.8 & 72.8     & 46.6 & 36.7   & 47.3    & 84.9  & 77.4    & 34.1     & 61.4    \\
O1 Embedder(Qwen2.5-7B)     & 86.0       & 65.8 & 72.5     & 45.4 & 37.0   & 46.3    & 86.9  & 75.8    & 35.5     & 61.2    \\
O1 Embedder(Qwen2.5-3B)     & 85.9       & 64.7 & 70.4     & 45.4 & 36.0   & 45.5    & 85.1  & 75.0    & 35.1     & 60.3    \\
O1 Embedder(Qwen2.5-1.5B)   & 84.3       & 61.9 & 68.3     & 43.1 & 35.8   & 43.7    & 83.0  & 74.3    & 34.0     & 58.7    \\
O1 Embedder(Qwen2.5-0.5B)   & 85.0       & 56.0 & 61.7     & 38.6 & 33.0   & 39.7    & 81.5  & 68.9    & 33.1     & 55.4    \\ \hline
\end{tabular}
}

\caption{Detailed scores of different base model for zero-shot evaluation in section~\ref{subsection: Robustness}.}
\label{tab:detailed scores}
\end{table*}

\section{Algorithm}

\begin{algorithm}
\caption{Data Production Process}
\label{alg:data_production}
\begin{algorithmic}[1]
\Require  
  \Statex Query-document dataset $D = \{(q_i, d_i)\}_N$ 
  \Statex Teacher model for generating thoughts $LLM$
  \Statex Retrieval models $R = \{r_1, r_2, ..., r_{|R|}\}$ 
  \Statex Example count $m$, candidate thoughts $k$ 
  \Statex Prompt tamplate: \textsc{Prompt}, Instruction: $Ins$
  \Statex Functions: 
  \begin{itemize}
    \item \textsc{SampleExamples}: Samples $m$ examples from $D$
    \item $\sigma^{r}$: Similarity score function of Retrieval model $r$
    \item \textsc{Voting}: majority voting function
  \end{itemize}

\Ensure  
  \Statex Enhanced dataset $\hat{D} = \{(q_i, t_i, d_i)\}_N$

\State Initialize $\hat{D} \gets \emptyset$
\For{each $(q_i, d_i) \in D$}
    \For{$j=1$ to $k$}
        \State $E \gets \textsc{SampleExamples}(D)$
        \State $P \gets \textsc{Prompt}.format(Ins,E,q_j) $ 
        \State $t_j \gets LLM.generate(P)$
    \EndFor
    \For{$j=1$ to $|R|$}
        \State $t^*_r \gets \textsc{argmax}(\{\sigma^r(t_j, d)\}_{j=1...k})$
    \EndFor
    \State $t_i \gets \textsc{Voting}(t^*_r)_{r \in R}$
    \State $\hat{D} \gets \hat{D} \cup \{(q_i, t_i, d_i)\}$
\EndFor
\State \Return $\hat{D}$
\end{algorithmic}
\end{algorithm}

The Algorithm ~\ref{alg:data_production} outlined in the Data production process in section~\ref{section: data production}. For each query-document pair \((q_i, d_i)\) in the dataset, the algorithm first samples \(m\) examples from \(D\) to create a prompt, which includes specific instructions. It then generates \(k\) candidate thoughts \(t_j\) by formatting the prompt with the sampled examples. Next, for each retrieval model \(r\), it calculates the similarity scores between the candidate thoughts and the document \(d_i\), selecting the thought with the highest score as \(t^*_r\). Finally, the algorithm aggregates these top thoughts through a majority voting mechanism to determine the final thought \(t_i\). The enhanced dataset \(\hat{D}\) is then constructed, consisting of the original queries, the generated thoughts, and their corresponding positive documents.

\section{Additional Case analyses \& Attention}
\label{section:additional_case_study}

\begin{table}[ht]
\begin{tabular}{p{0.9\linewidth}}

\hline
\textbf{Query:}   Hayden is a singer-songwriter from Canada, but where does Buck-Tick hail from?      \\
\hline
\textbf{Thought:}   \textcolor{softgreen}{\textbf{Buck-Tick}} is a \textcolor{softgreen}{\textbf{Japanese}} rock band, and its members are from various parts of \textcolor{softgreen}{\textbf{Japan}} and their music is a unique blend of alternative rock, gothic rock, and visual kei styles.   \\
\hline
\textbf{Positive Doc:}   Buck-Tick \textcolor{softgreen}{\textbf{Buck-Tick}} (stylized as BUCK-TICK) is a \textcolor{softgreen}{\textbf{Japanese}} rock band, formed in Fujioka, Gunma in 1983. The group has consisted of ...   \\
\hline
\end{tabular}
\caption{Another Example from HotPotQA (zero-shot).}
\label{tab:case3}
\end{table}

\begin{figure}[ht]
  \centering
  \includegraphics[width=0.9\linewidth]{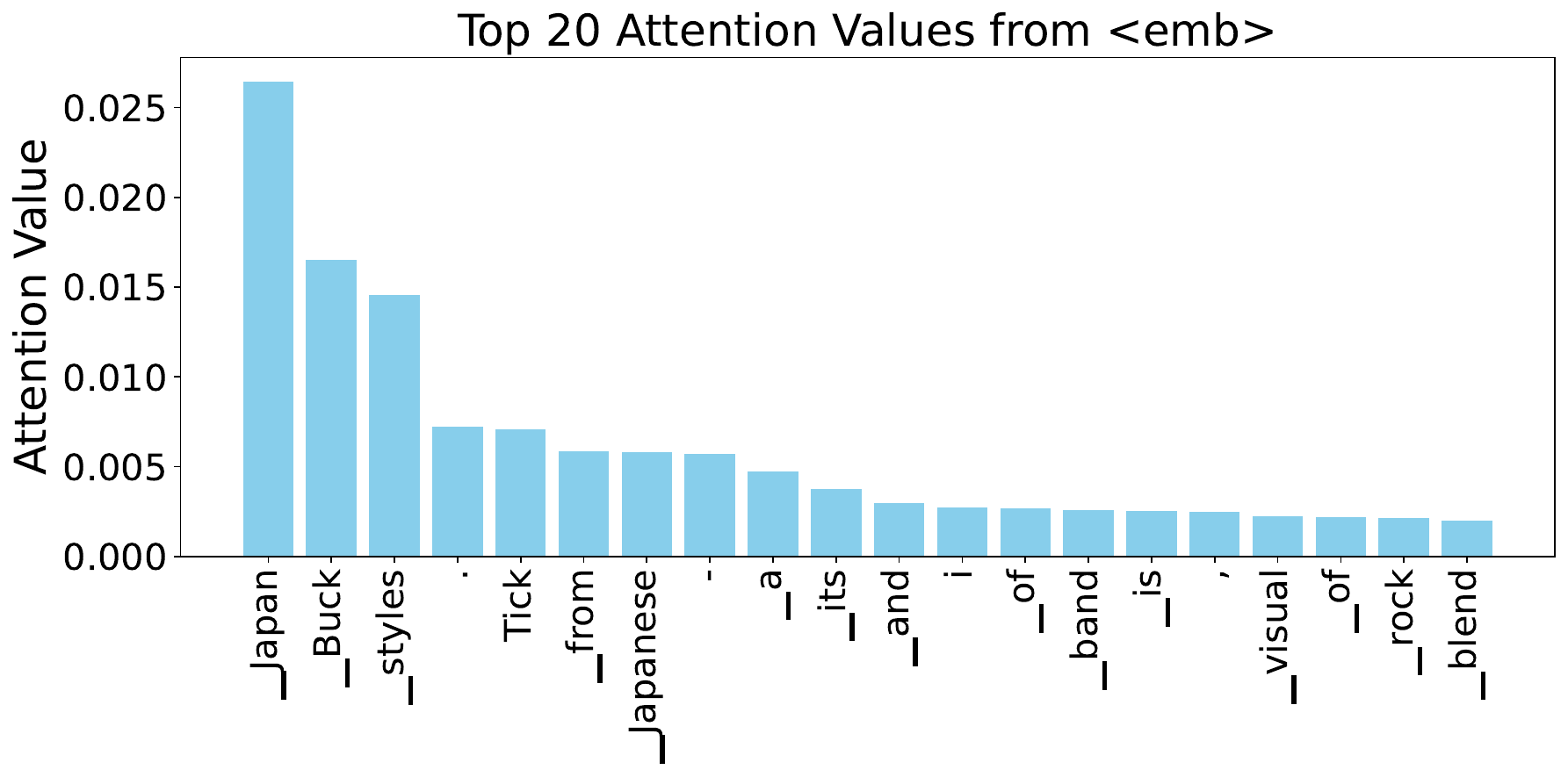}
  \caption{Top 20 Attention score from <emb> token in the thought}   
  \label{fig:top20}
\end{figure}

To better understand why thought can better help with retrieval, we further analyzed the attention scores of the model for the simple example in Table~\ref{tab:case3}. We calculate the attention weight of the <emb> token with each token $t_j$ in the thought:
$$
emb\_att(t_j)=\sum_{i} \frac{\exp{(q_{<emb>}^{i}\cdot k_{j}^{i}})}{\sum_{l=1}^{|L|} \exp{(q_{<emb>}^{i}\cdot k_{l}^{i})}}
$$
where $q_{<emb>}^{i}$ and $k_{j}^{i}$ are the "query" vector and the $j$th token's "key" vector of the $i$th attention head. $|L|$ is the total length of input sequence.

To enhance the clarity of our findings, we present the top 20 attention scores from the "<emb>" token in the generated thought, illustrated in Figure ~\ref{fig:top20}. The results reveal that the model assigns high scores to similar patterns found between the thought and the positive document, indicating its effectiveness in recognizing and focusing on relevant generated patterns. Notably, the word "Japan" emerges with the highest attention score, despite not appearing in the original query. This highlights the critical role of the generated thought in the retrieval process, as it enables the model to supplement essential information that may be absent from the query.



\section{Complex Retrieval Example}
\label{app: complex}

\begin{table*}[ht]
\begin{tabular}{p{0.9\linewidth}}
\toprule 
\textbf{Query:}   python check if a directory is writable      \\
\hline
\textbf{Thought:} \newline
To check if a directory is writable in Python, you can use the os.access() function. This function takes two arguments: the path to the directory and the mode you want to check (e.g. os.W\_OK for writing). If the directory is writable, the function returns True, otherwise it returns False.\newline Here's an example:
\begin{verbatim}
import os
path = '/path/to/directory'
if os.access(path, os.W_OK):
    print('The directory is writable')
else:
    print('The directory is not writable')
\end{verbatim} 
Alternatively, you can also use the os.path.isdir() function to check if the path is a directory, and then use the os.access() function to check the write permissions:
\begin{verbatim}
import os
path = '/path/to/directory'
if os.path.isdir(path) and os.access(path, os.W_OK):
    print('The directory is writable')
else:
    print('The directory is not writable')
\end{verbatim} 
\\ \hline
\textbf{Positive Doc:}
\begin{verbatim}
def _writable_dir(path):
    """
    Whether `path` is a directory, to which the user has write access.
    """
    return os.path.isdir(path) and os.access(path, os.W_OK)
\end{verbatim} 
\\ \hline
\end{tabular}
\caption{An example of O1 embedder solving a complex code retrieval problem.}
\label{tab:cosqa}
\end{table*}

We present another case from CosQA dataset demonstrating the capabilities of O1 embedder in Table~\ref{tab:cosqa}, to effectively generate and retrieve relevant code snippet. The query posed, "python check if a directory is writable" illustrates a common programming challenge. The O1 embedder responds by generating a comprehensive thought that not only provides a direct solution using the os.access() function but also includes an example code snippet. This case highlights the model's ability to synthesize information and present it in an accessible format, thereby aiding more accurate query representation. Additionally, the thought generated by the model incorporates alternative methods, such as verifying if the path is indeed a directory before checking write permissions. This demonstrates the O1 embedder's depth of understanding by providing multiple approaches to the problem. 
Through this case, we illustrate that the O1 embedder is capable of producing coherent, contextually relevant outputs that significantly improve the utility of code retrieval systems.

\end{document}